\documentclass{article}
\usepackage{graphicx} 
\usepackage{amsmath}
\usepackage[top=2.5cm,bottom=2.5cm,left=2.5cm,right=2.5cm]{geometry}
\usepackage[section]{placeins}
\usepackage{float}
\usepackage{geometry}
\usepackage{longtable}
\usepackage{gensymb}
\usepackage{amssymb}
\usepackage{amsfonts}
\usepackage{comment}
\usepackage{pdflscape}

\title{Marker-free eye-gaze estimation using a single image and depth from defocus}

\author{
David Hurtubise-Martin,
Feriel Fass,
Djemel Ziou\thanks{Corresponding author},
Marie-Flavie Auclair-Fortier
\\[0.6ex]
\small Département d'informatique, Faculté des sciences,
Université de Sherbrooke, Québec, Canada
}

\date{}

\begin{document}

\maketitle

\begingroup
\renewcommand{\thefootnote}{}
\footnotetext{%
\small\raggedright
\textit{Email addresses:}
David.Hurtubise-Martin@USherbrooke.ca (David Hurtubise-Martin),
Feriel.Fass@USherbrooke.ca (Feriel Fass),
Djemel.Ziou@USherbrooke.ca (Djemel Ziou),
Marie-Flavie.Auclair-Fortier@USherbrooke.ca (Marie-Flavie Auclair-Fortier)
\par
}
\endgroup

\begin{abstract}

\emph{This paper presents a marker-free eye-gaze estimation approach using a single 2D camera, such as an integrated laptop webcam. The gaze-related features are estimated from iris localization and head pose estimated by using depth from defocus. A variational Bayesian multinomial logistic regression (VBMLR) framework is used as mapping from the estimated features to the position of regard, based on an 8-dimensional feature vector of head-pose and iris-displacement parameters. No external marker is needed. Experiments were conducted by estimating the gaze of people watching a computer screen at different distances and compared against five existing methods. The obtained scores demonstrate the effectiveness of the proposed approach.
}


\end{abstract}

\section{Introduction}

Eye-gaze estimation is an active area of research in computer vision, cognitive sciences, marketing, and safe driving. In cognitive sciences, it is used for the study of neurological disorders, cognitive states, attention allocation, and human behaviours~\cite{marketing,Underwood2009,Underwood1998v,Duchowski}. In marketing, recommendations of goods and services can be personalized to a user based on what they are watching on the computer screen~\cite{Deshpande2025}. For safety, it is useful for monitoring the vigilance of vehicle drivers~\cite{geriatrics5020036,kapitaniak2015application}. Furthermore, in the fields of cinematography, eye-gaze estimation may be utilized to automate virtual camera control, optimize real-time shot selection, and analyze viewer attention for data-driven video editing~\cite{Achary2024, Treuting2006, Rachavarapu2019}.
	
Both active and passive computer-vision-based approaches have been proposed for eye-gaze estimation~\cite{Kar17,survey,illum1,por2,por3,por4}. For example, there are approaches in which electronic devices such as cameras equipped with active-illumination systems are used~\cite{zhang2024high}, and others with infrared illumination or complex and expensive systems~\cite{electronics10243165,por3,por4}. Some of these approaches deal with head-motion limitations~\cite{vidhya2025real}, and some require calibration procedures or multiple cameras~\cite{wang2025accurate,svm1}. Most of the proposed approaches use passive color cameras operating in the visible spectrum or 3D cameras for the relief estimation of the human face~\cite{jafari2015eye,svm1,jafari2012eye}. 
In the context of this work, we are interested in the case where the computer screen is the scene watched by users in front of a computer equipped with an embedded camera. In other words, given an image of the user's face taken by the embedded camera, we estimate the position on the screen that they are looking at. The main idea is to estimate suitable image features and then map these features to a position on the screen. These features include the position and orientation of the head, eye location, and a reference point detected automatically on the user's face. In the proposed marker-free model, the head-related features are estimated from depth-related cues inferred from blur in the image. The mapping is seen as a discriminative learning problem implemented using a variational Bayesian multinomial logistic regression (VBMLR) framework~\cite{Ziou2014}. Note that, in the case of discriminative learning, it has been pointed out that logistic regression (LR) outperforms the support vector machine (SVM)~\cite{svm1}. The paper is organized as follows. Section~2 reviews related work on single RGB camera-based gaze estimation. Section~3 presents an overview of the proposed system. Section~4 describes the VBMLR mapping estimation framework. Section~5 details the iris localization and geometric 
feature extraction. Section~6 presents the blur-based depth estimation pipeline. Section~7 reports the experimental protocol and results, and the conclusion is presented in Section~8.

\section{Related work}
In this section, we review representative works proposing eye-tracking models based on a single RGB camera embedded in handheld devices (laptops, smartphones, or tablets), without active illumination (e.g., infrared) or additional sensors. The compared works are summarized in Table~\ref{tab:gaze-survey}, in which the related work is organized according to the following criteria: (i) the underlying approach type (e.g., geometry-based, appearance-based, or hybrid), (ii) the predicted target (e.g., 2D screen point of regard (PoR) or gaze direction), (iii) the model and inference strategy, (iv) the preprocessing pipeline and extracted features, (v) the dataset characteristics, including size, number of participants, and acquisition conditions, (vi) the calibration requirements, and (vii) the reported performance metrics.

A first fundamental distinction concerns the predicted target. Some methods are designed to estimate the point of regard (PoR) on a screen (desktop, tablet, or smartphone), which is directly aligned with human–computer interaction applications. This formulation is common in webcam and mobile-based systems \cite{Falch2024,Huang2015,Krafka2016,Papoutsaki2016}. In contrast, other methods aim to estimate the gaze direction, typically represented as yaw and pitch angles or as a gaze vector in a normalized camera coordinate system. These methods dominate the core computer vision literature \cite{Deng2017,Kellnhofer2019,Zhang2017,Zhang2020}. In addition, some video-based methods incorporate temporal information to refine gaze estimates over time or to model predictive uncertainty, rather than producing a single instantaneous output \cite{Kellnhofer2019,Park2020}.

The earliest monocular gaze estimation methods primarily rely on geometric modeling. They estimate eye contours, iris shape, limbus geometry, or head pose, and subsequently infer gaze direction either analytically or through lightweight regression models \cite{Chen2008,Valenti2009}. While such methods are interpretable and frugal, their underlying assumptions often break down in real-world conditions, particularly under low image quality, the presence of glasses, large head pose variations, and unconstrained user behavior.
Appearance-based methods emerged as an alternative by replacing explicit geometric modeling with direct regression from image descriptors. They are generally simpler and can perform well in controlled, and device-specific settings. They rely heavily on handcrafted feature engineering and tend to generalize poorly to in-the-wild variability. A representative example is TabletGaze, which uses Histogram of Oriented Gradients (HoG) features combined with Random Forest regression to predict PoR on tablets \cite{Huang2015}. 
Extending this paradigm, interaction-based methods such as WebGazer leverage user interactions (e.g., mouse clicks) as implicit supervision to learn a mapping from webcam-based eye appearance to PoR coordinates \cite{Papoutsaki2016,Papoutsaki2017}. Although this enables scalable and easily deployable solutions, the learned mapping becomes tightly coupled to the specific device, user, screen configuration, and interaction context.
More recently, deep appearance-based methods rely on learning end-to-end representations directly from raw images. They predict either the PoR or the gaze direction, depending on the dataset and target application \cite{Krafka2016,Zhang2017}. They typically achieve state-of-the-art performance when trained on large-scale datasets and supported by robust normalization pipelines \cite{Cheng2021,Zhang2017,Zhang2018}. However, despite these advances, they remain sensitive to domain shift, exhibit person-specific biases, often require calibration, and suffer from inconsistencies in evaluation metrics across datasets \cite{Cheng2021,Zhang2019}.
The iTracker system demonstrated that large-scale consumer datasets can enable accurate, real-time PoR estimation on mobile devices \cite{Krafka2016}. Building on this direction, MPIIGaze and subsequent work shifted the focus toward person-independent gaze direction estimation in unconstrained laptop environments \cite{Zhang2017}.
Recent developments have further improved performance by incorporating structural and contextual information into deep learning frameworks. Some methods integrate full-face context to better capture head–eye interactions \cite{10.1145/3688636.3688650}, while others explicitly model geometric constraints within the learning process \cite{Deng2017}. More recently, transformer-based architectures have been introduced, primarily modifying feature fusion mechanisms rather than redefining the prediction objective. In practice, hybrid CNN–transformer models generally outperform pure transformer architectures, while neural architecture search (NAS)-based approaches aim to reduce computational cost by replacing large backbones with more efficient designs \cite{Cheng2021,Nagpure2023}.

Across these three categories, preprocessing is not a secondary design choice; it fundamentally affects both accuracy and comparability \cite{Cheng2021}. Common steps include face and eye detection and cropping. In deep appearance-based methods, preprocessing typically becomes more involved, incorporating image normalization to reduce variability in head pose and camera distance, coordinate transformations between PoR labels and gaze direction representations, and the selection of input modalities (full face, both eyes, or a single eye) \cite{Cheng2021,Zhang2018}.

Calibration strategies vary substantially across methods, largely depending on the prediction target and modeling assumptions. Models that estimate gaze direction, typically represented as a gaze vector or yaw–pitch angles, are more portable across users and devices because predictions are expressed in a camera or normalized coordinate system. Through normalization procedures and training on large in-the-wild datasets, these models learn to absorb variability in head pose, illumination, and appearance, enabling generalization to unseen users without explicit per-user calibration \cite{Zhang2015MPIIGaze,Nagpure2023,Kuric2025RAGEnet}. In contrast, methods that estimate the PoR on a screen are intrinsically dependent on user-specific factors, including screen geometry, viewing distance, and individual biases. As a result, they often require calibration to align model outputs with the physical display. Mobile PoR-based approaches may incorporate explicit multi-point calibration to improve accuracy \cite{Huang2015,Krafka2016}. Similarly, geometric or hybrid approaches that explicitly model the mapping between gaze direction and screen coordinates generally require an initial calibration phase (e.g., a small set of screen points) \cite{Chen2008,Valenti2009,Falch2024}.
Finally, interaction-based systems rely on continuous self-calibration from user behavior \cite{Papoutsaki2016}.
These differences in calibration requirements are closely linked to dataset design and scale. Deep appearance-based methods require large-scale annotated datasets to ensure robustness to appearance, head pose, and lighting variations. Representative examples include MPIIGaze \cite{Zhang2015MPIIGaze} (over 200,000 images), and ETH-XGaze \cite{Falch2024} (1 million images) and GazeCapture \cite{Krafka2016} (approximately 2.45 million images). In such cases, geometric relationships are implicitly learned from data rather than explicitly modeled. In contrast, geometric projection approaches reduce dependence on large datasets by leveraging explicit modeling and calibration. However, validation on a single participant, as reported in \cite{Falch2024}, may limit statistical generalizability.

To evaluate these methods, evaluation metrics are chosen according to the predicted target.
For PoR estimation, performance is measured in screen-space units (pixels, millimeters, or centimeters), typically using metrics such as mean error (ME) or root mean square error (RMSE), which quantify the distance between predicted and ground-truth gaze points \cite{Krafka2016,Cheng2021}. However, these metrics are inherently dependent on the display and viewing setup. Pixel error varies with screen resolution, while physical units (mm or cm) depend on screen size and user–screen distance \cite{Cheng2021,Zhang2019}. Some studies convert screen-space error into visual angle to improve comparability, but this requires precise knowledge of screen geometry and viewing distance, which is not always reported \cite{Cheng2021}. In addition, some methods adopt a classification-based evaluation. For example, in \cite{Valenti2009}, the screen is divided into discrete areas of regard (AoR), and performance is measured by the accuracy of assigning each frame to the correct area. Consequently, direct comparison between PoR-based methods remains challenging. For gaze direction estimation, the standard metric is the mean angular error (in degrees), defined as the average angular deviation between predicted and ground-truth gaze directions, typically ranging from approximately 3° to 14° depending on the dataset and evaluation protocol \cite{Chen2008,Zhang2015MPIIGaze,Nagpure2023,Kuric2025RAGEnet}. However, these values are not directly comparable, as the same angular error can correspond to different screen-space errors, primarily depending on the user’s distance from the screen.
Moreover, some deep learning methods report computational efficiency metrics, such as the number of parameters and the number of operations. For instance, \cite{Nagpure2023} reports a model with 1.027 million parameters and 0.28 giga floating-point operations (GFLOPs), whereas \cite{Kuric2025RAGEnet} reports 28.66 million parameters and 9.79 GFLOPs.

Unlike most existing methods, which either learn gaze implicitly from image appearance using large annotated datasets or project gaze onto the screen through a separate geometric stage \cite{Krafka2016,Zhang2017,Falch2024}, the proposed method explicitly incorporates user-to-screen distance estimated using depth from defocus into the gaze-classification feature representation. The user-to-screen distance is estimated through 3D reconstruction from blur, enabling a hybrid formulation that combines appearance-based features with geometric information. The area of regard (AoR) on the screen is then predicted using Bayesian logistic regression based on a compact eight-dimensional feature vector. This approach aims to improve screen-region discrimination while maintaining low model complexity and high interpretability.

\newgeometry{
  left=5mm,
  right=2mm,
  top=15mm,
  bottom=15mm}
\begin{landscape}

\small

\begin{longtable}{|p{0.5cm}|p{1.8cm}|p{2cm}|p{3cm}|p{3.cm}|p{5cm}|p{1.5cm}|p{5.5cm}|}
\caption{Comparison of gaze estimation approaches}
\label{tab:gaze-survey} \\
\hline
\textbf{Ref} & \textbf{Type method} & \textbf{Predicted target} & \textbf{Model and inference} & \textbf{Preprocessing} & \textbf{Dataset} & \textbf{Calibration} & \textbf{Metrics and scores} 

\endfirsthead
\textbf{Ref} & \textbf{Approach type} & \textbf{Predicts} & \textbf{Model and inference} & \textbf{Preprocessing} & \textbf{Dataset characteristics (size, subjects, conditions)} & \textbf{Calibration} & \textbf{Dataset and score} \\
\endhead

\endfoot
\endlastfoot

\hline
\cite{Chen2008} 
& Geometry-based 
& Gaze direction 
& Single-camera geometric model 
& Eye localization, head pose 
& One subject; controlled lab setup; fixed head pose (50 cm) 
& Required (9 points)
& No movement: \newline $X_{acc} = 17.7$ mm (1.83\degree), \newline $Y_{acc} = 19.3$ mm (2.0\degree) \newline
With movement: \newline $X_{acc} = 22.42$ mm (2.18\degree), \newline $Y_{acc} = 26.17$ mm (2.53\degree). \\ 
\hline

\cite{Valenti2009} 
& Geometry-based 
& PoR 
& Geometric webcam model 
& Eye detection, geometric features 
& 20 subjects performing a webpage browsing task; constrained motion 
& Required 
& Accuracy = 95\% for AoR localization \\ 
\hline

\cite{Huang2015} 
& Appearance-based 
& PoR
& HoG + Random Forest 
& Eye-region extraction 
& TabletGaze: 41 participants; $\sim$100k images; tablet front camera; fixed distance; indoor lighting 
& Optional
& With calibration: ME = 2.5 cm, angular error = 2.86°–4.76° \newline 
Without calibration: ME = 3.17 cm, angular error = 3.63°–6.03° \\ 
\hline

\cite{Papoutsaki2016} 
& Interaction-based 
& PoR (pixels); visual angle (lab)
& Linear R.\footnote{Linear regression} (120D eye feature vector); RR\footnote{Ridge Regression} (RR); RR+C \footnote{Ridge Regression + cursor}; RR+C+F \footnote{Ridge Regression + cursor + fixation buffer}
& Face, eye, and pupil detection 
& WebGazer: 82 participants; 20,251 clicks; in-the-wild browser study; webcam RGB stream \newline
In-lab study: 4 participants; 962 clicks 
& Self-calibration 
& Mean error: \newline 174.9 px (remote RR+C); \newline 169 px (lab); \newline Visual angle: 4.17° \\ 
\hline

\cite{Krafka2016} 
& Deep appearance-based 
& PoR
& iTracker: end-to-end CNN for feature extraction + SVR for PoR regression 
& Face detection, eye cropping, face grid encoding, input normalization (224×224)
& GazeCapture: 1474 participants; $\sim$2.5M images; mobile devices; diverse environments; varying pose and lighting 
& Optional (0–13 points) 
& Mean screen error: \newline 
No calibration: 1.71 cm (mobile); 2.53 cm (tablet) \newline
With calibration (13 points): 1.34 cm (mobile); 2.12 cm (tablet) \\ 
\hline

\cite{Zhang2015MPIIGaze}
& Deep appearance-based 
& Gaze direction (yaw, pitch), 3D head pose
& Multimodal CNN (eye images + head pose); LeNet-style; L2 loss 
& Face detection, facial landmarks, eye cropping, normalization into canonical space, histogram equalization
& MPIIGaze: 15 participants; 213,659 eye images; laptop RGB camera; collected over $\sim$3 months; large illumination variability \newline
EYEDIAP: 16 participants; controlled lab setting 
& None 
& Mean angular error: \newline 13.9° (MPIIGaze); \newline 10.5° (EYEDIAP) \\ 
\hline

\cite{Nagpure2023} 
& Efficient hybrid deep 
& Gaze direction 
& NAS-based multi-resolution feature extractor + fusion transformer
& Multi-resolution facial feature extraction
& Benchmarks: MPIIGaze, GazeCapture, MPIIFaceGaze, Gaze360, RT-GENE, EYEDIAP; diverse conditions 
& None 
& 3.96° (MPIIFaceGaze), 10.52° (Gaze360), 6.40° (RT-GENE), 5.00° (EYEDIAP); \newline 
1.027M parameters; 0.28 GFLOPs\footnote{\label{Gflops}Giga Floating Point Operations} \\ 
\hline

\cite{Falch2024}
& Hybrid (appearance + geometry) 
& PoR via 3D user position estimation 
& CNN (OpenVINO / ETH-XGaze) + geometric projection + regression + SfM
& Appearance features + 3D reconstruction 
& Single-user dataset; RGB webcam (~30Hz); 2560×1440 screen; 800 mm viewing distance; validated with Tobii tracker 
& 4-point calibration 
& RMSE: \newline 
No head movement: 50 mm / 3.2° (ETH-XGaze), 53 mm / 3.3° (OpenVINO) \newline 
Yaw rotation: 80 mm / 5.1° \newline
Lateral translation: 60 mm / 4° \\ 
\hline

\cite{Kuric2025RAGEnet} 
& Appearance-based 
& Gaze direction (yaw, pitch) 
& Residual Attention-Based CNN (RAGE-Net); L2 loss 
& Eye cropping, normalization
& MPIIGaze: 15 participants; 213k images \newline
MPIIFaceGaze: 15 participants; $\sim$45k images; laptop RGB camera; indoor setup 
& None 
& Mean angular error: \newline 
3.96° (MPIIGaze); \newline 
4.08° (MPIIFaceGaze) \newline
28.66M parameters; 9.79 GFLOPs\footnote{\ref{Gflops}} \\ 
\hline
\end{longtable}
\end{landscape}
\restoregeometry

\section{Proposed approach overview}

Let us consider that a user is watching a document displayed on the screen of a computer equipped with an embedded 2D camera operating in the visible spectrum. The purpose is to predict the location within the document that the user is watching. The document content referenced in the screen space can be a text, a word, a picture, a drawing, among many others. The smallest element (i.e., resolution) that can be successfully predicted depends on the features of the human visual system, the features of the camera, the user location relative to the screen, the illumination, and the gaze estimation algorithms~\cite{survey,Valenti2009}. That is why the screen is logically subdivided into areas of a size making it possible to measure the change of the gaze. These areas, named areas of regard ($AoR$), are what we would like to predict. The assumptions behind the proposed approach are: 1) the user environment is illuminated, by both ambient light and the computer screen itself; 2) the area in the image occupied by the user's head allows the detection of points of interest; 3) the orientation of the user's head and the direction of his/her gaze are close. These assumptions are realistic because users often watch the screen in an illuminated environment. The third assumption is supported by the fact that, in intentional motion, there is usually a small latency between head movements and eye movements~\cite{Hu19}.

\begin{figure}[!ht]
    \centering
    \fbox{\includegraphics[width=0.95\textwidth]{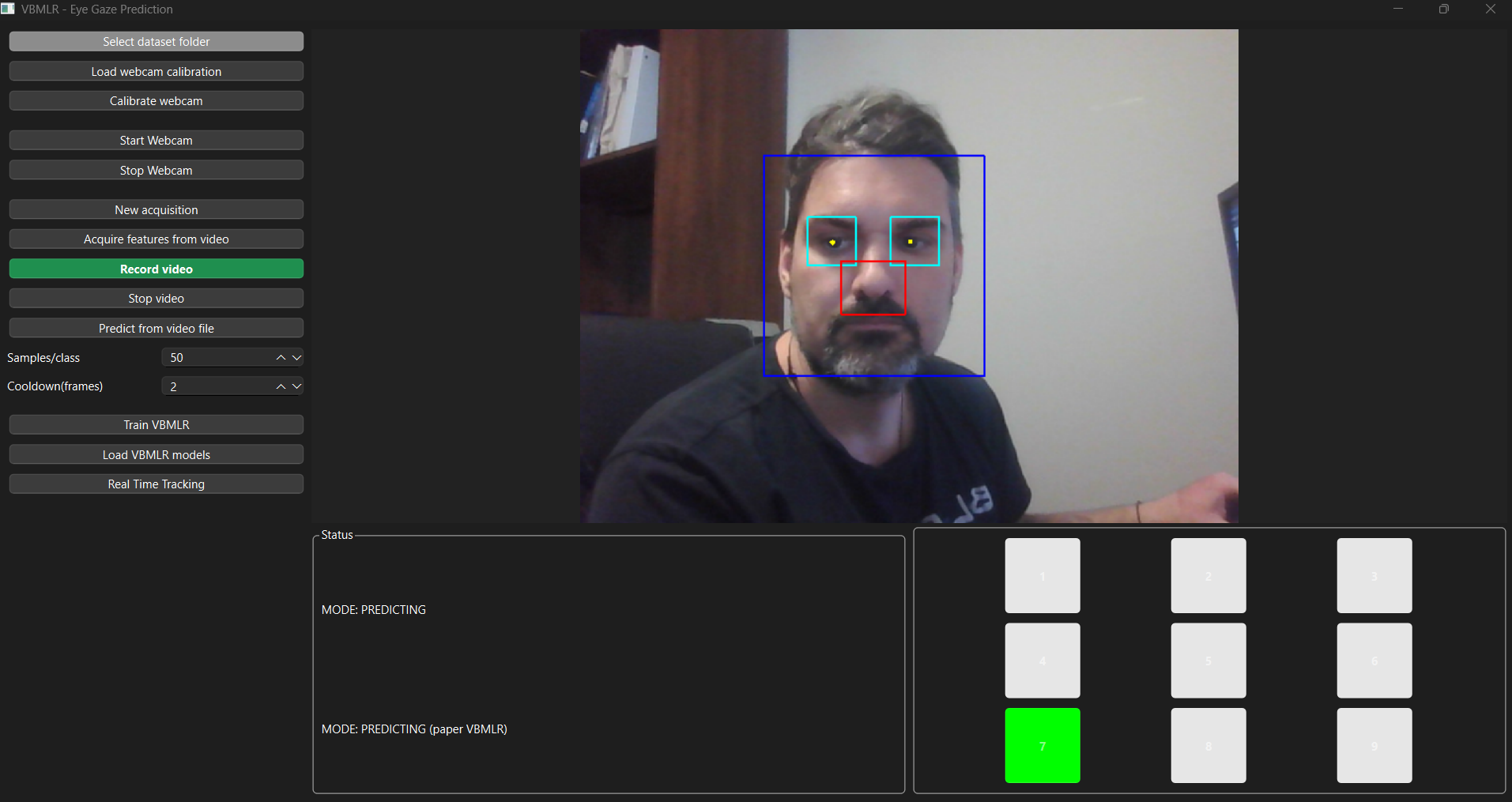}}
    \caption{Overview of the proposed marker-free gaze prediction system and real-time interface.}
    \label{fig:interface}
\end{figure}

The gaze prediction process includes image acquisition, feature extraction, and mapping. The architecture of the proposed system is illustrated through the interface shown in Fig.~\ref{fig:interface}. This interface displays the live camera feed with overlaid detection markers for the face and eyes, indicated by the blue and cyan bounding boxes, respectively. As the camera feed is processed, the area of regard watched by the user is predicted and highlighted in real time. For example, in Fig.~\ref{fig:interface}, the screen is divided into a discrete $3 \times 3$ grid, and the predicted $AoR$ is displayed in green. The mapping function associates an $AoR$ with a feature vector $d$ characterizing the iris displacement and the head pose. Formally, this relationship is defined as $AoR=f(d\mid\theta)$, where $\theta$ denotes the parameters of the mapping model. The mapping can be viewed as a prediction problem implemented using different types of predictors. It can be formulated as a continuous prediction, or regression, problem, in which a predictor estimates continuous screen coordinates or gaze directions. CNNs, for example, can learn nonlinear mappings from eye and face images to screen positions or gaze directions~\cite{Zhang2015MPIIGaze,Krafka2016}. Some approaches explicitly combine eye appearance and head-related information to improve gaze prediction~\cite{10.1145/3688636.3688650,Hu19}. Other continuous prediction models use residual neural networks~\cite{Li19} or Transformer-based architectures~\cite{ChengLu2021}.

Alternatively, the mapping can be formulated as a classification problem, since each $AoR$ can be considered as a class. It therefore consists in assigning an $AoR$, i.e., a class, to an observation vector $d$. Several methods can be used to perform this classification, including neural networks, SVM, and LR models~\cite{nair2010rectified,cortes1995support,svm1,svm2}. In the present work, the mapping is built using a classification model in which the classes correspond to the $AoR$s. Let us consider labelled spatial data $s=(s_1,\cdots,s_M)$, where $s_i$ is a binary variable indicating whether the $AoR_i$ is watched by the user, and $M$ is the number of $AoR$s. The mapping model is first learned from labelled data and then used for real-time prediction at each frame. More precisely, given an image acquired at time $t$, a feature vector $d$ is first extracted and assigned to one of the predefined $AoR$ classes.

\begin{figure}[H]
    \centering
    \fbox{\includegraphics[width=0.95\textwidth]{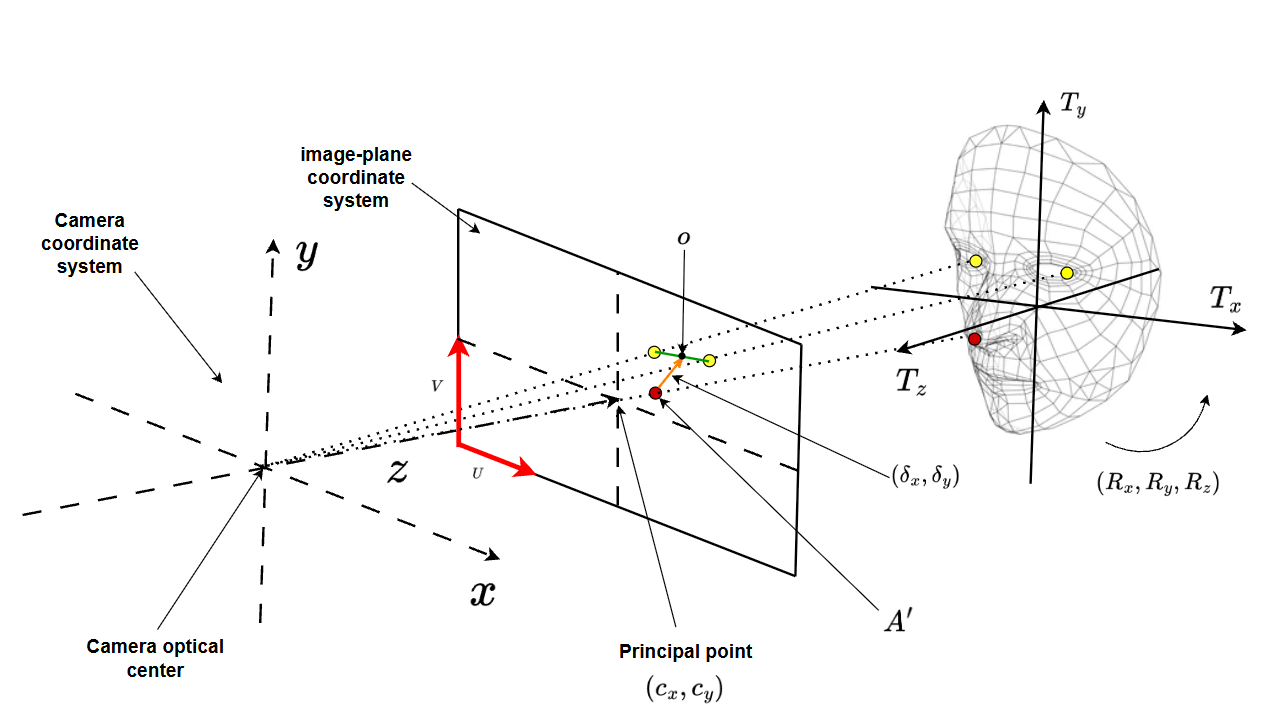}}
    \caption{Representation of the coordinate systems and gaze
    parameters used in the proposed method. In the image-plane
    coordinate system, the iris displacement vector
    $\delta=(\delta_x,\delta_y)$ is defined from the nose center $A'$ to
    the mean iris position $o$. In the camera coordinate system, the
    head-position vector $T=(T_x,T_y,T_z)$ and the head-orientation
    vector $R=(R_x,R_y,R_z)$ describe the position and orientation of
    the head relative to the camera optical center.}
    \label{fig:coord}
\end{figure}

Let us consider two coordinate systems. In the image-plane coordinate
system $(u,v)$, a point is represented by its horizontal and vertical
pixel coordinates. The principal point $(c_x,c_y)$ corresponds to the
intersection of the camera optical axis with the image plane and is
located approximately at the center of the image. In the three-dimensional camera coordinate system, the origin is located
at the camera optical center. A three-dimensional point is represented
by its coordinates $(x,y,z)$. The feature space is formed by eight gaze parameters. The iris displacement vector $\delta=(\delta_x,\delta_y)$ is expressed
in the image-plane coordinate system. It is defined as the displacement
from the nose center $A'$ to the mean iris position $o$, where both
points are detected and represented by their pixel coordinates in the
image. The head-position
vector $T=(T_x,T_y,T_z)$ and the head-orientation vector
$R=(R_x,R_y,R_z)$ are expressed in the three-dimensional camera
coordinate system. Figure~\ref{fig:coord} illustrates these coordinate
systems and the gaze parameters expressed within them.

In the current marker-free implementation, the gaze features are
estimated directly from the acquired facial image. The iris positions
and the nose center are detected in the image and used to compute the
iris displacement vector. Facial reference points are also detected and
combined with blur-based depth information to estimate the head position
and orientation without requiring physical markers attached to the
face. Temporal stabilization is applied to the iris localization and
depth estimation to improve reliability during real-time acquisition.

Logistic regression is a well-established method which can outperform the SVM in related discriminative settings~\cite{svm1,svm2}. However, logistic regression may not yield finite parameter estimates when the classes are separable. A Bayesian formulation addresses this limitation by imposing prior distributions on the model parameters. Variational Bayesian logistic regression (VBLR)~\cite{svm2} approximates the posterior distribution of these parameters and provides a tractable framework for Bayesian binary classification. However, the present gaze prediction problem involves several $AoR$ classes. A multinomial formulation is therefore required. In this work, the variational Bayesian multinomial logistic regression (VBMLR) framework~\cite{Ziou2014} is used to estimate $p(s\mid d,\theta)$, where $\theta$ is the parameter vector. The main idea is to compare each in-screen class $i$ with the outside-screen baseline class ($i=0$). This baseline-based formulation is used for training and prediction. A one-versus-one strategy is also used in the comparative mapping evaluation for multiclass benchmarking under the same eight-dimensional feature representation.

To summarize, the screen is virtually divided into areas of regard ($AoR$s). Given an image, the feature vector $d=(R_x,R_y,R_z,T_x,T_y,T_z,\delta_x,\delta_y)$ is estimated and mapped to an $AoR$. This mapping is formulated as a classification problem using VBMLR. The mapping model is learned from labelled data.

\section{Mapping estimation}

Let us recall that the feature vector is $d=[R_x, R_y, R_z, T_x, T_y, T_z, \delta_x, \delta_y]$. Let $i \in \{1,\dots,M\}$ be the in-screen $AoRs$, and we reserve $i=0$ for the outside-screen baseline class. Each class $i$ is associated with a binary variable $s_i$, and at a given time, the user watches exactly one class, either one $AoR$ or the outside class; that is, $\sum_{i=0}^{M} s_i = 1$. Our objective is to design a model which enables us to predict the area watched by the user at a given time; i.e., find $i$ such that $s_i=1$. To estimate the value of \(s_i\), we first need to estimate its probability and then determine \(s_i\) using a Bayesian decision rule. 
More formally, let us consider the set of complete data
$\mathcal{D}=\{(\Omega_{0},s_{0}),\dots,(\Omega_{M},
s_{M}) \}$, where $\Omega_i$ contains the $n_i$ feature vectors $d_{i,1}$ to 
$d_{i,n_i}$ of the $i^{th}$ area. Set $\Omega_0$ of vectors when the user watches outside the screen, is the baseline class. 
Let us recall that the mapping is considered as a classification problem and  carried out by using Bayesian multinomial LR. Experiments in related works show that the performance of the VBLR outperforms many well established discrimination based algorithms such as the SVM, Relevance Vector Machine, Bayesian Logistic Regression Model, Informative Vector Machine, and  LR~\cite{svm2}. Since the gaze-estimation problem involves multiple classes, the VBLR formulation is extended to the multinomial case, referred to as VBMLR~\cite{Ziou2014}. The multinomial
logistic regression consists in estimating the statistical model $\theta=(\theta_1, \dots, \theta_{M})$ that discriminates
each class $i \in [1,M]$ from the baseline
class~\cite{Begg84,Krishnapuram}.
The accuracy of VBMLR-based classification may decrease when
either the relative size of the baseline class decreases or the number of classes increases, or the classes are unbalanced \cite{Ziou2014, ben2024prediction}. The computational time increases with the number of
classes and make it difficult to fulfil the real-time  requirement. Fortunately, the number of AoR is not high, e.g.  2x2 inches give around 25 areas for 15 inches screen.

Let us assume that $\Omega_0$ and $\Omega_i$ are generated from  known probability density functions (pdfs) $q_0(d)$ and $q_i(d)$ respectively. Let us recall that each feature vector $d$ of $\Omega_i$ (resp. $\Omega_0$) is assigned with two binary variables $s_i=1$ and $s_0=0$ (resp. $s_i=0$ and $s_0=1$) if it is generated by the $i^{th}$ area model (resp. outside model). We assume that the sets $\Omega_i$ are balanced. The probability of the binary variables is given by $p(s_{i}=1|d,\theta_i)=\frac{\mathrm{exp}(\theta_{i}^{T}d)}{\sum_{j=1}^{M}\mathrm{exp}(\theta_{j}^{T}d)}$, and $p(s_{i}=0|d,\theta_i)=1-p(s_{i}=1|d,\theta_i)$. The maximum likelihood is a popular estimator of the parameter vector $\theta_i$. However, collinearity, separability, and lack of parsimony are potential drawbacks. To address these limitations, Bayesian formulations have been proposed for both binary classification \cite{svm2} and multinomial classification \cite{Ziou2014}. Let $p(\theta_i)$ be a prior; we need to find the parameter vector
$\theta_i$ maximizing the posterior probability
$p(\theta_i|s_0=0,s_i=1)$ given by:
\begin{equation}max_{\theta_i} p(\theta_i|s_0=0,s_i=1)~\propto ~p(\theta_i) \sum_{d \in \Omega_i} \prod_{k\in \{0,i\}} p(s_k=g(k)|d,\theta_i)q_k(d), \label{affectrule}\end{equation}
with
\begin{equation}
g(k)=
\left\{
\begin{array}{ll}
0, & \text{if } k=0,\\
1, & \text{otherwise}.
\end{array}
\right.
\end{equation}

However, in the case of multidimensional data, the estimation of $\theta_i$ fails due to the insufficiency of computer accuracy when computing the exponential function $p(s_i=1|d,\theta_i)$ of~Eq.~\eqref{affectrule}. Fortunately, variational approximation
and Jensen's inequality can be used to approximate the posterior:
\begin{equation}
p(\theta_i|s_0=0,s_i=1)  \propto p(\theta_i) \prod_{k \in
\{0,i\}} F(\epsilon_k) e^{(E_{q_k}(H_k)-\epsilon_k)/2 -
\varphi(\epsilon_k) (E_{q_k}(H_k^2)-\epsilon_k^2)},
 \label{affectrule1}
\end{equation}
where $H_k=(2g(k)-1) \theta_k^T d_k$, $E_{q_k}$ is the expectation with respect to $q_k$, $\varphi(\epsilon_k) = tanh(\epsilon_k/2)/4\epsilon_k$, and $\epsilon_k$ is a variational parameter. Let us consider that the prior $p(\theta_i)$ is a Gaussian with mean $\mu_i$ and covariance $\Sigma_i$, the approximation of the posterior above is a Gaussian with a posterior mean $\mu_i^{post}$ and a posterior covariance
$\Sigma_i^{post}$ given by
\begin{eqnarray}
&&(\Sigma_i^{post})^{-1} = (\Sigma_i)^{-1} + 2 \sum_{k \in \{0,i\}}
\varphi(\epsilon_k) E_{q_k}(d_k d_k^T)\\
\nonumber &&\mu_i^{post} = \Sigma_i^{post}(\Sigma_i^{-1} \mu_i
+\sum_{k \in \{0,i\}}(g(k)-0.5)E_{q_k}(d_k)) \nonumber \\
&& \epsilon_k^2= E_{q_k}(d_k^T \Sigma_i^{post} d_k) + (\mu_i^{post})^t
E_{q_k}(d_k^T  d_k) \mu_i^{post},\quad k \in \{0,i\} \nonumber.
 \label{Stat}
\end{eqnarray}
Mathematical derivation of the above formulae and other details
can be found in~\cite{svm2}. These equations are not devoted to the estimation of the vector parameter $\theta_i$. However, this vector can be set to $\mu_i^{post}$ because $\theta_i$ follows a Gaussian distribution with mean $\mu_i^{post}$. The implementation of the mapping learning based gaze estimation is straightforward. 
The learning phase consists of estimating $\mu_i^{post}$ (i.e. $\theta_i$) using Eq \ref{Stat}. The detailed algorithm of VBMLR is given in \cite{svm2}. For the test step, given a feature vector $d$, the $i^{th}$ area is the  watched one if
\begin{equation}
i =
\begin{cases}
\displaystyle \operatorname*{arg\,max}_k \;
p(s_k=1 \mid d,\theta_k)\,p(\theta_k)\,q_k(d),
& \text{if} \quad
p( s_i=1 \mid d,\theta_i )p(\theta_i)q_i(d) > \eta, \\[6pt]
0, & \text{otherwise}.
\end{cases}
\label{Detection}
\end{equation}
The threshold $\eta$ can be estimated through cross-validation during the learning phase by minimizing false positives, minimizing false negatives, or maximizing the trade-off between them.

\section{Iris localization and geometric feature extraction}

The goal is to estimate the feature vector
$d=(R_x,R_y,R_z,T_x,T_y,T_z,\delta_x,\delta_y)$.
The iris centers, the nose center, and the displacement
$\delta=(\delta_x,\delta_y)$ are expressed in image coordinates, where
a point is represented by its horizontal and vertical coordinates
$(u,v)$. The head orientation $R=(R_x,R_y,R_z)$ and the head position
$T=(T_x,T_y,T_z)$ are expressed in the three-dimensional camera frame
$(x,y,z)$, whose origin is located at the camera center.

The iris centers are first localized in the image. However, their
absolute image coordinates depend both on eye motion and on the global
position of the face in the image. For example, if the user moves the
head to the left while maintaining the same gaze direction, the detected
iris centers also move to the left, even though their positions relative
to the face remain approximately unchanged. A facial reference point is
therefore used to express the iris position relative to the face. The
center of the detected nose region is selected because it is located
near the central facial region and moves with the head. Let
$A'=(a_x,a_y)$ denote the center of the nose region in image coordinates.

After the center of each iris has been estimated and smoothed over
consecutive frames, the two resulting centers are averaged to obtain the
mean iris position $o=(o_x,o_y)$. The vector from the nose center to this
mean iris position is defined as
\begin{equation}
\delta
=
(\delta_x,\delta_y)
=
o-A'.
\label{eq:iris_displacement}
\end{equation}

\begin{figure}[H]
    \centering
    \fbox{\includegraphics[width=0.78\textwidth]
    {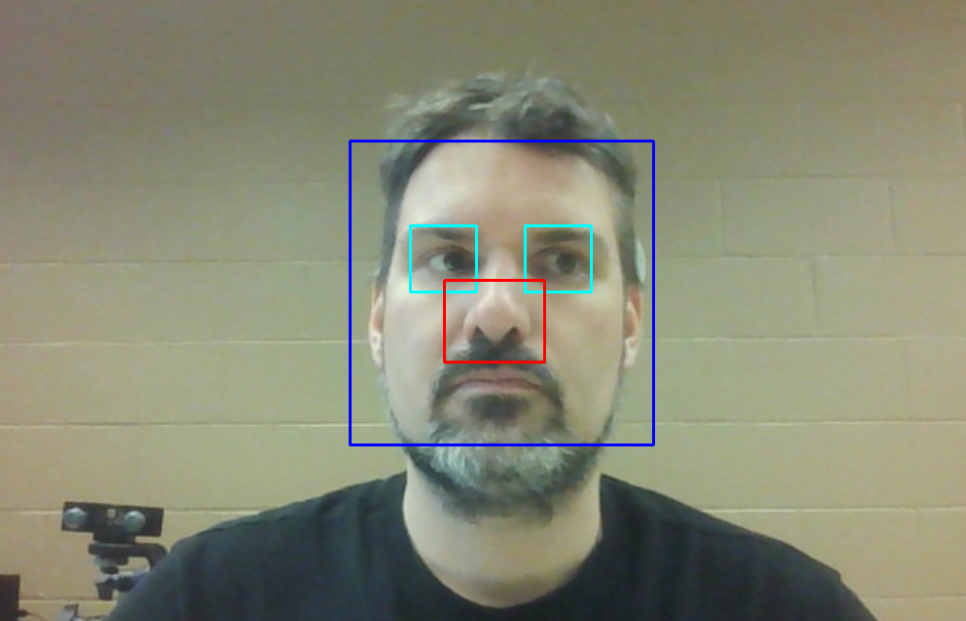}}
    \caption{Example of the detected face region, the eye regions derived
    from the face bounding box, and the detected nose region.}
    \label{fig:bounding_box}
\end{figure}

To estimate $o$ and $A'$, the face and nose regions are localized using
Viola--Jones~\cite{Viola01}. Once the face bounding box has been
detected, the two eye regions are placed inside its upper part according
to the expected anatomical positions of the eyes. Their size and
position are defined relative to the detected face box, so that the eye
regions automatically become larger or smaller when the apparent face
size changes in the image. Figure~\ref{fig:bounding_box} shows an
example of the detected face, the resulting eye regions, and the
detected nose region. The iris center is then estimated independently
within each eye region.

Each estimated center is smoothed by combining its current position
with the position retained for the same eye in the preceding frame.
When both resulting iris centers are available, their horizontal and
vertical coordinates are averaged. The resulting point, located midway
between the two iris centers, defines the mean iris position $o$.

More precisely, let us first assume that the eyes are open. The iris
boundary can be modeled as a circular step edge. It is possible to
estimate the iris center by formulating the problem as an optimization
based on the radial derivative of the scale-space Radon
transform~\cite{Daugman93}. However, webcam images are often noisy, the
iris contour may be incomplete or partially occluded by the eyelids, and
an exhaustive search over all possible center and radius values is not
well suited to a real-time implementation.

Instead, we use a faster approximation in which we first detect a
plausible dark iris region, then verify the presence of a visible iris
contour, use the contour geometry to obtain a more precise estimate of
the iris center, and finally smooth this estimated position over
consecutive frames to reduce small variations caused by noise or
temporary occlusion.

\begin{figure}[!ht]
    \centering
    \fbox{\includegraphics[width=0.95\textwidth]
    {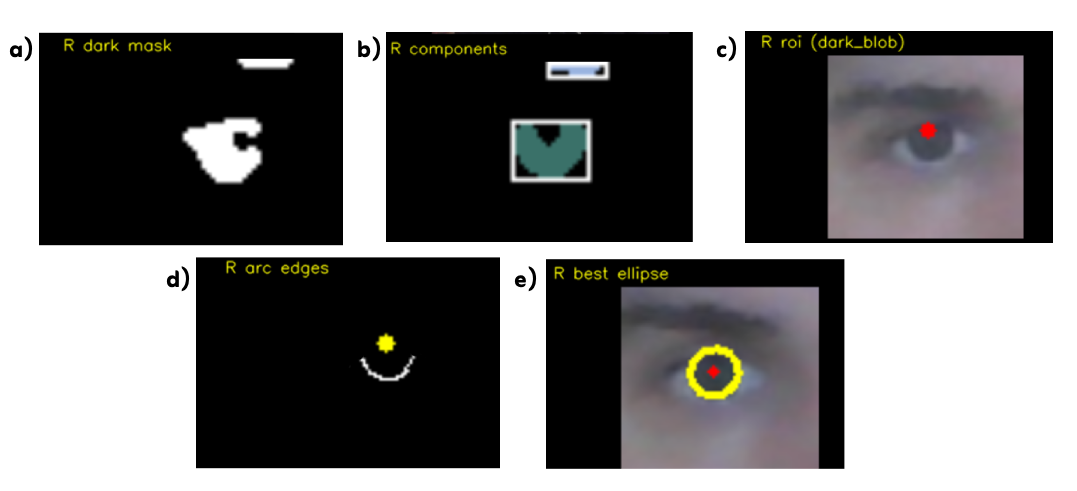}}
    \caption{Main steps of iris-center estimation in the right eye.
    (a) Binary dark mask computed in the eye ROI.
    (b) Separate dark regions obtained from the mask.
    (c) Initial iris-center candidate obtained from the selected dark
    region.
    (d) Contour verification around the candidate center.
    (e) Refined iris-center measurement obtained from the fitted
    ellipse.}
    \label{fig:iris_pipeline}
\end{figure}

The first practical step is therefore to restrict the search to the
darkest part of the eye ROI, because the iris and pupil usually form one
of the darkest structures in that region. A binary dark mask is computed
from low-intensity pixels and then morphologically cleaned. This mask is
shown in Fig.~\ref{fig:iris_pipeline}(a).

In the second step, the dark mask is divided into separate dark regions,
and the most plausible one is selected as the iris candidate. The center
of this retained region provides an initial estimate of the iris center
for the eye currently being processed. For the contour evaluation below,
this initial candidate is denoted by $(\hat{o}_x,\hat{o}_y)$.
Figure~\ref{fig:iris_pipeline}(b) shows the different dark regions
obtained from the mask, and Fig.~\ref{fig:iris_pipeline}(c) shows the
initial center obtained from the selected dark region.

At this stage, we have a plausible estimate of the iris center, but it
is not yet certain that this is the true center of the iris because the
visible iris region may be partially truncated. The next step is
therefore to verify whether this candidate is surrounded by a visible
contour compatible with the iris. The visible contour portions extracted
around the candidate center are illustrated by the white arc in
Fig.~\ref{fig:iris_pipeline}(d). These contour portions determine the
sampled directions that are retained for the following circular-contour
evaluation.

\begin{figure}[H]
    \centering
    \fbox{\includegraphics[width=0.40\textwidth]
    {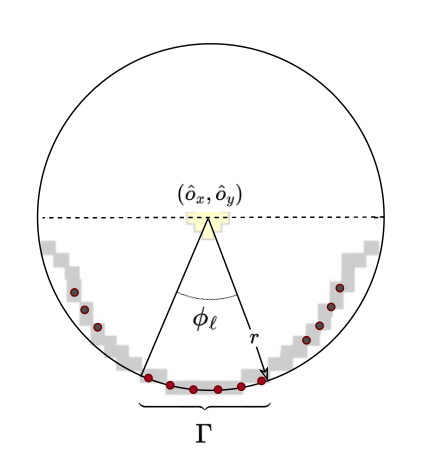}}
    \caption{Circular-contour evaluation around the initial iris-center
    candidate $(\hat{o}_x,\hat{o}_y)$. The angle $\phi_\ell$ defines a
    sampled direction and $r$ denotes the tested radius. The retained
    directions $\Gamma$ are represented by their intersections with the
    tested circular contour along the visible iris-contour portion.}
    \label{fig:iris_contour_sampling}
\end{figure}

To evaluate this visible contour, $N_\phi$ directions are uniformly
sampled on a tested circular contour of radius $r$ centered at
$(\hat{o}_x,\hat{o}_y)$. The sampled directions are defined by
\begin{equation}
\phi_\ell
=
\frac{2\pi\ell}{N_\phi},
\qquad
\ell=0,\ldots,N_\phi-1.
\label{eq:iris_sampling_angle}
\end{equation}
Among these directions, only those corresponding to the visible contour
portions are retained. Let $\Gamma$ denote the set of retained sampled
directions. Figure~\ref{fig:iris_contour_sampling} illustrates the
candidate center, a tested radius $r$, a sampled direction
$\phi_\ell$, and the retained directions $\Gamma$.

For a tested radius $r$, the corresponding contour score is defined as
\begin{equation}
L_{\mathrm{iris}}(\hat{o}_x,\hat{o}_y,r)
=
\frac{1}{|\Gamma|}
\sum_{\ell\in\Gamma}
I\!\left(
\hat{o}_x-r\sin\phi_\ell,\;
\hat{o}_y+r\cos\phi_\ell
\right),
\label{eq:centeriris_lineint}
\end{equation}
where $|\Gamma|$ denotes the number of retained sampled directions.
Thus, $L_{\mathrm{iris}}$ represents the mean image intensity evaluated
along the visible portion of the tested circular contour. The
normalization by $|\Gamma|$ makes this score independent of the number
of retained directions. The radial edge evidence is then measured using the same retained
sampled directions $\Gamma$ on two neighbouring circular contours:
\begin{equation}
\Delta L(\hat{o}_x,\hat{o}_y,r)
=
L_{\mathrm{iris}}(\hat{o}_x,\hat{o}_y,r)
-
L_{\mathrm{iris}}(\hat{o}_x,\hat{o}_y,r-1),
\label{eq:centeriris_radial_diff}
\end{equation}
which measures the change in mean image intensity between two
neighbouring circular contours.

For the fixed candidate center $(\hat{o}_x,\hat{o}_y)$, this comparison
is repeated for the candidate radii between $r_{\min}$ and $r_{\max}$.
For each radius, the corresponding radial intensity difference is
computed. The resulting sequence is locally smoothed to reduce small
variations caused by image noise, and the radius associated with the
largest magnitude of the radial intensity change is retained. A strong
radial intensity change indicates a likely transition between the iris
and its surrounding region and therefore provides evidence of an iris
boundary compatible with the tested circular contour.

Once a plausible iris contour has been identified, an ellipse is fitted
to the available contour points following the ellipse-based iris
localization approach proposed in~\cite{Plopski2016}. Let
$j\in\{\mathrm{right},\mathrm{left}\}$ identify the eye currently being
processed. The center of the fitted ellipse provides the current
iris-center measurement $\tilde{o}_t^{\,j}$.
Figure~\ref{fig:iris_pipeline}(e) shows the retained contour, the fitted
ellipse, and its center.

The measured iris center may vary slightly from one frame to the next
because of image noise, illumination changes, partial eyelid occlusion,
or variations in the detected contour. To reduce these variations, the
current measurement is combined with the center retained for the same
eye in the previous frame. The resulting iris-center position is defined
as
\begin{equation}
o_t^{\,j}
=
\lambda_t^{\,j}\,o_{t-1}^{\,j}
+
\left(1-\lambda_t^{\,j}\right)\tilde{o}_t^{\,j},
\label{eq:centeriris_ema}
\end{equation}
where $\lambda_t^{\,j}\in[0,1]$ controls the balance between the center
retained in the preceding frame and the current measurement for the same
eye. The confidence is derived from the iris-detection score produced by
the selected dark region or by the fitted ellipse. A low-confidence
measurement increases $\lambda_t^{\,j}$, so that the resulting position
remains closer to the preceding center. A high-confidence measurement
decreases $\lambda_t^{\,j}$, so that the resulting position follows the
current measurement more closely.

This operation is performed independently for the two eyes. When both
resulting iris centers are available, their mean position is
\begin{equation}
o
=
\frac{
o_t^{\mathrm{right}}
+
o_t^{\mathrm{left}}
}{2}.
\label{eq:mean_iris_position}
\end{equation}

\begin{figure}[!h]
    \centering
    \fbox{\includegraphics[width=0.72\textwidth]
    {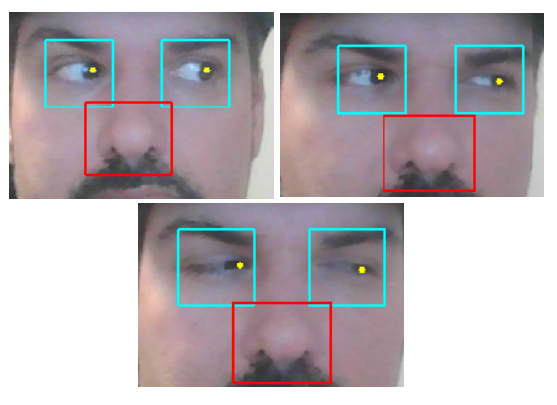}}
    \caption{Examples of marker-free face, eye, and nose localization
    with iris-center estimation for open and partially closed eyes.}
    \label{fig:iris_localization_example}
\end{figure}

In practice, this procedure allows the iris center to remain localized
under moderately challenging conditions, including partial occlusion by
the eyelids and short blinking phases. As illustrated in
Fig.~\ref{fig:iris_localization_example}, the iris center remains
localized across different eye states.

\section{Blur-based depth estimation}	

Performing the mapping requires the estimation of the feature vector
$d$. Both binocular and monocular vision methods can be used to estimate the geometric information required for this purpose~\cite{survey, Deng2017, Chen2008}. In this work, we
propose to exploit image blur acquired with a single 2D camera. The estimation of $d$ involves three main steps: (i) estimation of the
image blur, (ii) estimation of the depths of facial reference points,
and (iii) estimation of the head-position and head-orientation
components of the feature vector. We begin with the blur-estimation
stage.

During image acquisition, the scene radiance falling on the image plane is corrupted mainly by blur caused by the camera. Let us consider the corruption model as a convolution of the light $I(u,v)$ falling on the lens, the Gaussian point spread function (PSF) $g(u,v;\varrho)$ of the lens, where $\varrho$ is the blur measure. The light coming out of the lens is given by:
\begin{equation}
 I(u,v; \varrho_b) = I(u,v) * g(u,v;\varrho),
 \label{eq:imform_blur}
\end{equation}
where $*$ denotes convolution. The blur $\varrho$ depends on the $z$ coordinate of the 3D point $(x,y,z)$ described in the camera coordinate system. The image point $(u,v,f)$ is the perspective projection of $(x,y,z)$. Several PSFs $g(\cdot)$ have been proposed in the literature; among them, the Gaussian filter is one of the most widely used~\cite{psf1}. The relationship between depth and blur is given by~\cite{blur1,blur2}:
\begin{equation}
  z = \left\{ \begin{array}{rl}
\frac{Ff}{f-F-kF_N \varrho} & \mbox{if } z \ge u_f \\
\frac{Ff}{f-F+kF_N \varrho} & \mbox{otherwise},
\end{array} \right.
\label{eq:dist_z} 				
\end{equation}
where $u_f$ is the distance between the lens and the position of perfect focus, $F_N$ is the f-number, $f$ is the distance from the lens to the image plane, and $F$ is the focal length. According to Eq.~\eqref{eq:dist_z}, the computation of the depth $z$ requires the estimation of the blur measure $\varrho$ and the camera
parameters $F$, $f$, $k$, and $F_N$. These camera parameters are
obtained from the camera specifications or estimated through
calibration. In our implementation, the camera calibration is performed
using the \texttt{calibrateCamera} function available in OpenCV. The
blur measure $\varrho$ is estimated using the method described
in~\cite{blur1}.

A simple and fast strategy is used to estimate the local blur from the
change in image gradients produced by an additional Gaussian smoothing.
Let $I(u,v;\varrho_b)$ denote the acquired image. A second image is
obtained by applying a Gaussian filter of known scale $\varrho_0$:
\begin{equation}
I(u,v;\varrho_1)
=
I(u,v;\varrho_b)
*
g(u,v;\varrho_0),
\label{eq:imform_blur_smooth}
\end{equation}
where $\varrho_1$ denotes the effective blur level of the additionally
smoothed image. Because the additional Gaussian smoothing reduces local
intensity variations, the gradient magnitude of
$I(u,v;\varrho_1)$ is generally smaller than that of
$I(u,v;\varrho_b)$. Inspired by the spatial-domain principle described
in~\cite{blur1}, the unknown blur is approximated as
\begin{equation}
\varrho(u,v)
=
\varrho_0
\sqrt{
\frac{
\left\|\nabla I(u,v;\varrho_b)\right\|
}{
\left\|\nabla I(u,v;\varrho_1)\right\|
}
-1
}.
\label{eq:gradient_ratio_blur}
\end{equation}
where $\nabla$ denotes the image gradient. In the implementation, the
input color image is first converted to grayscale. Its local contrast is
then adjusted using contrast-limited adaptive histogram equalization
(CLAHE), which performs histogram equalization within small image
regions while limiting excessive noise amplification. The additional
Gaussian filter is subsequently applied, and the horizontal and vertical
gradients of the acquired and additionally smoothed images are computed
using discrete derivative filters. These derivatives measure local
intensity changes in the horizontal and vertical directions. The resulting values define a local blur map $\varrho(u,v)$ over the
image. For each selected facial reference point, the blur value is
obtained by sampling this map within a small local neighbourhood. The
sampled blur value is then used together with the camera parameters to
compute the corresponding depth $z$ using
Eq.~\eqref{eq:dist_z}.

\begin{figure}[!h]
    \centering
    \fbox{\includegraphics[width=0.92\textwidth]{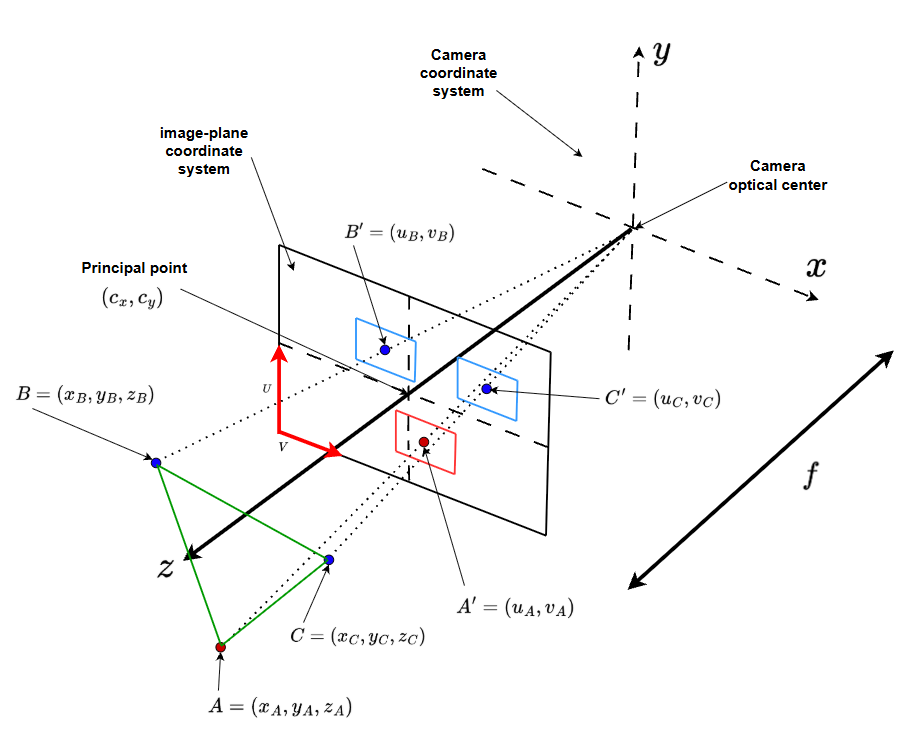}}
    \caption{Geometry used for head pose estimation. The image points $A'$, $B'$,
and $C'$ are extracted from the detected facial regions and assigned depths
$z_A$, $z_B$, and $z_C$ estimated from blur. These depths are then used to
calculate the 3D points
$A$, $B$, and $C$.}
    \label{fig:face_geo}
\end{figure}

Let us now consider the estimation of the head pose in the 3D camera coordinate system.
The objective of the following steps is to define three facial reference
points in the image-plane coordinate system $(u,v)$, reconstruct their
corresponding three-dimensional in the camera coordinate system $(x,y,z)$, and then estimate the head orientation vector
$R=(R_x,R_y,R_z)$ and the head position vector
$T=(T_x,T_y,T_z)$. A representation of the coordinate systems and
information used for this estimation is shown in
Fig.~\ref{fig:face_geo}.

Let $A'=(u_A,v_A)$, $B'=(u_B,v_B)$, and $C'=(u_C,v_C)$ denote three facial reference points expressed in the image-plane coordinate system $(u,v)$. The point $A'$ is defined as the center of the detected nose bounding box, while $B'$ and $C'$ are defined as the centers of the detected eye bounding boxes. The local blur values $\varrho_A$, $\varrho_B$, and $\varrho_C$ are sampled in neighbourhoods centred at these points. Using Eq.~\eqref{eq:dist_z}, these blur values are converted into the corresponding depths $z_A$, $z_B$, and $z_C$. Each depth is the third component of the three-dimensional coordinates $(x,y,z)$ of the corresponding facial point in the camera coordinate
system and represents its position along the optical direction of the camera.

The depth value alone does not determine the complete three-dimensional
position of a facial point. It must be combined with its image
coordinates and the intrinsic parameters of the camera. More precisely,
for an image point $(u,v)$ with depth $z$, the corresponding coordinates
$(x,y,z)$ in the camera coordinate
system are obtained by back-projection:
\begin{equation}
x=\frac{z(u-c_x)}{f_x},
\qquad
y=\frac{z(v-c_y)}{f_y},
\label{eq:corresp}
\end{equation}
where $(c_x,c_y)$ denotes the principal point of the camera, corresponding to the pixel coordinates at the center of the image, and $f_x$ and $f_y$ are the focal lengths expressed in pixels. Therefore, the image coordinates
$(u,v)$ determine the direction of the projection ray, while the
estimated depth $z$ determines the position of the point along this ray.

Applying this back-projection independently to $A'$, $B'$, and $C'$
gives the three-dimensional facial points
\begin{equation}
\begin{aligned}
A &=
\left(
\frac{z_A(u_A-c_x)}{f_x},
\frac{z_A(v_A-c_y)}{f_y},
z_A
\right),\\
B &=
\left(
\frac{z_B(u_B-c_x)}{f_x},
\frac{z_B(v_B-c_y)}{f_y},
z_B
\right),\\
C &=
\left(
\frac{z_C(u_C-c_x)}{f_x},
\frac{z_C(v_C-c_y)}{f_y},
z_C
\right).
\end{aligned}
\label{eq:facial_points_3d}
\end{equation}
The reconstructed points $A$, $B$, and $C$ define a triangular geometric configuration in the camera coordinate system. The head orientation is estimated as the normalized normal vector of the facial plane defined by these three
points:
\begin{equation}
R=
\frac{(B-A)\times(C-A)}
{\left\|(B-A)\times(C-A)\right\|}.
\label{eq:head_orientation}
\end{equation}
The translational component is taken as the centroid of the three reconstructed facial points,
\begin{equation}
T=
\frac{A+B+C}{3}.
\label{eq:head_translation}
\end{equation}
which is then used in the feature vector together with $R$ and
$\delta$.

To summarize, we first estimate a local blur measure at the selected
facial reference points detected in the image and convert these
measurements into depth values. The estimated depths are then combined
with the image-plane coordinates of the detected points and the
intrinsic camera parameters to reconstruct the corresponding facial
points in the camera coordinate system. These reconstructed points are
finally used to estimate the head-orientation vector
$(R_x,R_y,R_z)$ and the head-position vector
$(T_x,T_y,T_z)$ without requiring any physical markers. Together with the iris displacement vector
$(\delta_x,\delta_y)$, the head-orientation, head-position, and iris
displacement vectors form the eight-dimensional feature vector $d$ used
for gaze estimation.

\section{Experimentations}

\begin{figure}[!htbp]
    \centering
    \fbox{\includegraphics[width=0.72\textwidth]{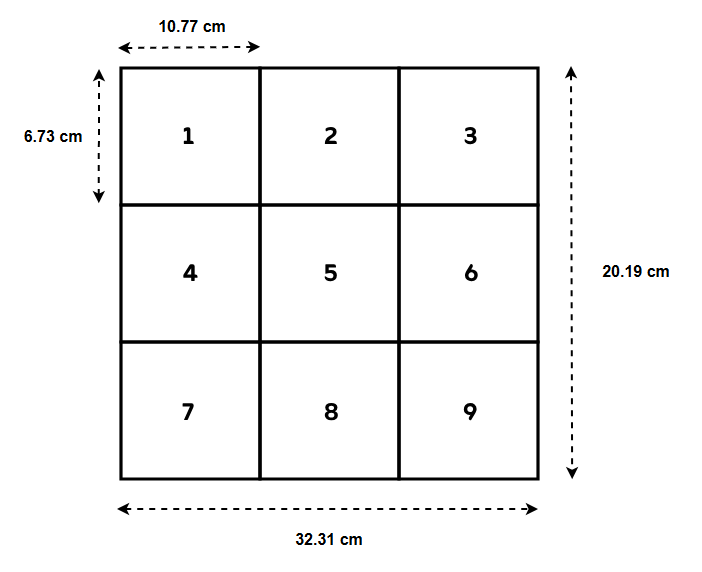}}
    \caption{15 inches screen subdivision.}
    \label{fig:screen_subdiv}
\end{figure}

Experiments were conducted using the integrated webcam of an MSI Cyborg 15~A12V laptop.
The webcam stream is processed at a resolution of $640\times480$ RGB pixels.
Camera intrinsic parameters were estimated beforehand using the OpenCV camera calibration
procedure and then used in the geometric back-projection and blur-to-depth estimation stages
described in Eqs.~\eqref{eq:corresp} and~\eqref{eq:dist_z}. A 15 inches screen was divided into nine zones as shown in Fig.~\ref{fig:screen_subdiv}, each
of 10.77 cm x 6.73 cm. The dataset was acquired with 4 subjects. For each subject, 9 areas of regard (\(AoR\)s)
were recorded with 100 images per \(AoR\), yielding 900 images per subject.
The same protocol was repeated at two user-to-camera distances: 50~cm and 70~cm.
The complete dataset therefore contains \(4 \times 9 \times 100 \times 2 = 7200\) images.
For each distance-specific evaluation (50~cm or 70~cm), the aggregated data contain
\(4 \times 100 = 400\) samples per zone and 3600 samples in total per distance.
Under the 5-fold stratified cross-validation protocol, this corresponds, in each fold, to 320 training samples and 80 test samples per zone.
In the global pooled evaluation (50~cm and 70~cm), each zone contains 800 samples
(\(4\times100\times2\)), for a total of 7200 samples. In that case, each fold contains 640 training samples and 160 test samples per zone.

We compare the proposed variational Bayesian multinomial logistic regression (VBMLR), described in Eqs.~\eqref{affectrule}--\eqref{Detection}, with five representative baselines. The mapping is learned from the 8-dimensional feature vector $d=[R_x,R_y,R_z,T_x,T_y,T_z,\delta_x,\delta_y]$,
introduced in section~3. Since one bias term is added to them, each class uses \(8+1=9\) learned parameters. For the 9 screen zones, this corresponds to \(9\times(8+1)=81\) learned parameters. In this formulation, the hyperparameters \(\mu\) and \(\Sigma\) define the Gaussian prior in Eq.~\eqref{affectrule}. A Ridge regression mapping is included as representative of the WebGazer
mapping strategy~\cite{Papoutsaki2016}. SVM with RBF kernel and multiclass linear regression~\cite{Begg84, Papoutsaki2016} complete the comparison as strong classical baselines. A lightweight 1D convolutional neural network (CNN-1D) for low-dimensional structured features~\cite{vidhya2025real,Zhang2015MPIIGaze}, and a ResNet18
baseline initialized from ImageNet pretraining.
All mappings are trained and evaluated using a strict one-versus-one strategy for the
9 $AoR$ zones. In all methods, the same 8-dimensional feature vector and the same 5-fold stratified
cross-validation protocol are used. In addition, we report an image-based ResNet18 fine-tuned from ImageNet as a high-capacity reference.

For all mapping methods, the hyperparameters were selected through preliminary experiments conducted under the same cross-validation protocol. Table~\ref{tab:hyperparams_models} summarizes the main theoretical choices and optimization settings considered for each model, with the retained values shown in bold.

\begin{table}[H]
\caption{Hyperparameters for Each Mapping Method}
\label{tab:hyperparams_models}
\centering
\small
\setlength{\tabcolsep}{6pt}
\renewcommand{\arraystretch}{1.08}
\begin{tabular}{p{2.8cm} p{10.2cm}}
\hline
\textbf{Model} & \textbf{Parameters} \\
\hline

\textbf{VBMLR} &
\textit{prior mean \(\mu\)}: \textbf{0}, 0.3, 1, 10, 25, 50, 100 \\
&
\textit{prior covariance \(\sigma\)}: 0.3, 1, 10, 25, 50, \textbf{100} \\
&
\textit{stopping criterion}: $10^{-3}$, \textbf{\boldmath$10^{-4}$}, $10^{-5}$ \\
\hline

\textbf{SVM (RBF)} &
\textit{regularization parameter} $C$: \textbf{0.3}, 1, 10, 25, 50, 100 \\
&
\textit{kernel}: \textbf{RBF\footnotemark[1]} \\
\hline

\textbf{LR Multiclass} &
\textit{prediction rule}: argmax \\
\hline

\textbf{Ridge} &
\textit{regularization}: \textbf{L2} \\
&
\textit{regularization coefficient} $\alpha$: 0.3, 1, 10, 25, \textbf{50}, 100 \\
\hline

\textbf{CNN-1D} &
\textit{hidden layers}: (\textbf{8}, \textbf{16}) \\
&
\textit{activation}: \textbf{RelU\footnotemark[2]} \\
&
\textit{optimizer}: \textbf{Adam\footnotemark[3]} \\
&
\textit{dropout}: \textbf{0.3} \\
&
\textit{learning rate}: \textbf{\boldmath$10^{-3}$}, $10^{-4}$, $10^{-5}$ \\
&
\textit{iterations}: 10, \textbf{25}, 50 \\
\hline

\textbf{ResNet18} &
\textit{input representation}: \textbf{\boldmath$2\times4$ pseudo-image} \\
&
\textit{pretraining model}: \textbf{ImageNet\footnotemark[4]} \\
&
\textit{optimizer}: \textbf{Adam} \\
&
\textit{learning rate}: $10^{-3}$, \textbf{\boldmath$10^{-4}$}, $10^{-5}$ \\
&
\textit{iterations}: 5, \textbf{10}, 20 \\
\hline
\end{tabular}
\end{table}

\footnotetext[1]{Radial Basis Function (RBF) is a non-linear kernel used in SVMs to map data into higher-dimensional spaces~\cite{cortes1995support}.}

\footnotetext[2]{Rectified Linear Unit (ReLU) preserves information about relative intensities as it travels through multiple layers of feature detectors~\cite{nair2010rectified}.}

\footnotetext[3]{Adaptive Moment Estimation (Adam) is an algorithm for first-order gradient-based optimization of stochastic objective functions, based on adaptive estimates of lower-order moments~\cite{kingma2014adam}.}

\footnotetext[4]{ImageNet is a large-scale hierarchical database of images built upon the backbone of the WordNet structure~\cite{deng2009imagenet}.}

For VBMLR, each mapping is trained using a variational Bayesian logistic-regression formulation with a Gaussian prior of zero mean, \(\mu=\mathbf{0}\), and
covariance \(\Sigma=\sigma^2 I\), with \(\sigma=100\). The iterative procedure is stopped when the largest absolute change in the estimated posterior mean $\mu^{post}$ between two successive iterations becomes smaller than $10^{-4}$. In SVM, we use $C=0.3$ which denotes the SVM regularization parameter controlling the trade-off between maximizing the distance between the separating hyperplane and the nearest support vectors, and minimizing training errors. 
Ridge uses $\alpha=50$, where $\alpha$ denotes the $L_2$ regularization coefficient. For CNN-1D, we adopt a compact architecture tailored to the 8-dimensional
feature vector, with two convolutional layers followed by ReLU activations. The
first layer produces 8 feature channels and the second produces 16, followed by
a small pooled representation and a lightweight fully connected head with
dropout ($0.3$) to limit overfitting on low-dimensional inputs.
The network is trained with the Adam optimizer for 25 iterations with a learning rate of $10^{-3}$.
We also evaluated a pretrained ResNet18 model. The proposed 8-dimensional feature vector was reshaped into a small $2 \times 4$ pseudo-image, upsampled by bilinear interpolation to $224 \times 224$, and then replicated across color channels to form a pseudo-image compatible with the standard ResNet18 input format. ResNet18 is fine-tuned with the Adam optimizer for 10 iterations with a learning rate of $10^{-4}$, using ImageNet pretrained weights.

Tables~\ref{tab:metrics_50cm}--\ref{tab:metrics_global} summarize the performance and deployment cost of the compared mappings. The reported metrics include the mean classification accuracy in percentage (\%), the mean number of false positives (FP) and true negatives (TN) averaged over the 9 $AoR$ zones, the mean training time per test feature vector in milliseconds (ms), and the total number of learned parameters across all $AoR$s. We also report error statistics in percentage (\%), namely the minimum error, maximum error, and standard deviation of the classification error rate across the five cross-validation folds. ResNet18 is not included in the table because its use is difficult to justify other than by the obtained scores. Indeed, it learns at the pixel level and specializes based on nominal characteristics. Although these characteristics are calculated from the pixel, it is not easy to provide a model explaining the basis of fine-tuning in this case.

\begin{table}[H]
\caption{Distance-specific comparison at 50~cm using only non-pretrained models. Metrics are averaged over the 9 $AoR$ zones (400 samples per zone).}
\centering
\small
\resizebox{\textwidth}{!}{%
\begin{tabular}{lcccccccc}
\hline
Method & Mean Acc.\ (\%) & Mean FP & Mean TN & Training Time (ms) & Nb Params & Min Err.\ (\%) & Max Err.\ (\%) & Std Err.\ (\%) \\
\hline
VBMLR (proposed) & \textbf{99.69} & \textbf{1.22} & \textbf{3198.78} & 0.0018 & \textbf{81} & \textbf{0.14} & \textbf{0.83} & 0.27 \\
SVM (RBF)        & 99.03 & 3.89 & 3196.11 & 0.4498 & 35{,}073 & 0.28 & 1.53 & 0.42 \\
LR Multiclass    & 66.11 & 135.56 & 3064.44 & \textbf{0.0003} & \textbf{81} & 33.19 & 35.56 & 0.87 \\
Ridge            & 99.58 & 1.67 & 3198.33 & 0.0063 & 324 & \textbf{0.14} & 0.97 & 0.29 \\
CNN-1D           & 98.86 & 4.56 & 3195.44 & 0.0071 & 2{,}809 & 0.83 & 1.39 & \textbf{0.18} \\
\hline
\end{tabular}%
}
\label{tab:metrics_50cm}
\end{table}

At 50~cm, the proposed VBMLR reaches near-ceiling performance with 99.69\% mean accuracy in Table~\ref{tab:metrics_50cm}. Its fold-wise error statistics remain tightly bounded, with a minimum error of 0.14\%, a maximum error of 0.83\%, and a standard deviation of 0.27\%, indicating highly consistent generalization across folds. This behavior is confirmed by the confusion matrix in Table~\ref{tab:vbmlr_cm_50}, which is almost perfectly diagonal. The rare errors are isolated and mainly occur between spatially adjacent $AoR$s in the upper-left region. Ridge remains competitive at 99.58\%, followed by SVM at 99.03\% and CNN-1D at 98.86\%. Among the non-pretrained models, CNN-1D exhibits the lowest error variability, with a standard deviation of 0.18\%, whereas Linear Regression remains far below at 66.11\%, with large error values and high variability, confirming that a simple direct linear mapping is insufficient for a 9 zone nonlinear screen partition.

As an additional exploratory experiment, we also evaluated a pretrained ResNet18 model at 50~cm. Interestingly, it slightly exceeded VBMLR and reached 99.72\% mean accuracy, together with the lowest FP count (1.11) and the highest TN count (3198.89). This result is noteworthy because the ResNet did not receive a natural gaze image as input. Instead, the same handcrafted 8-dimensional feature vector was reshaped into a small pseudo-image, upsampled to the standard input resolution, and processed by an ImageNet-pretrained backbone. This suggests that transfer learning can still exploit discriminative structure even from such an artificial image-like representation. However, this slight gain must be interpreted with caution. The ResNet relies on external pretraining and uses 8.4M parameters, which implies a substantially higher overall computational cost than the non-pretrained mappings. The score difference with ResNet is not large enough to justify the much higher complexity, both in terms of the number of parameters and the amount of training data involved. The convolutional pipeline is a black box and the observed robustness cannot be attributed to explicit geometric cues, and the decision boundaries cannot be inspected or constrained. For this reason, it is discussed separately from the main comparison table.

\begin{table}[H]
\caption{Distance-specific comparison at 70~cm using only non-pretrained models. Metrics are averaged over the 9 $AoR$ zones (400 samples per zone).}
\centering
\small
\resizebox{\textwidth}{!}{%
\begin{tabular}{lcccccccc}
\hline
Method & Mean Acc.\ (\%) & Mean FP & Mean TN & Training Time (ms) & Nb Params & Min Err.\ (\%) & Max Err.\ (\%) & Std Err.\ (\%) \\
\hline
VBMLR (proposed) & \textbf{96.81} & \textbf{12.78} & \textbf{3187.22} & 0.0019 & \textbf{81} & 1.94 & \textbf{4.44} & 0.92 \\
SVM (RBF)        & 94.53 & 21.89 & 3178.11 & 0.4705 & 35{,}073 & 4.58 & 6.25 & \textbf{0.64} \\
LR Multiclass    & 63.42 & 146.33 & 3053.67 & \textbf{0.0003} & \textbf{81} & 35.14 & 38.19 & 0.97 \\
Ridge            & 93.50 & 23.89 & 3176.11 & 0.0063 & 324 & 5.42 & 7.64 & 0.92 \\
CNN-1D           & 96.64 & 14.33 & 3185.67 & 0.0071 & 2{,}809 & \textbf{1.81} & 5.42 & 1.22 \\
\hline
\end{tabular}%
}
\label{tab:metrics_70cm}
\end{table}

At 70~cm, all methods degrade due to the limitations of the camera's optical characteristics and therefore the degradation of depth due to defocusing. This degradation is visible in the fold-wise error statistics reported in Table~\ref{tab:metrics_70cm}, where the
non-pretrained models exhibit minimum errors ranging from 1.81\% to 35.14\%
and maximum errors ranging from 4.44\% to 38.19\%, depending on the method. Among the non-pretrained approaches, VBMLR remains the strongest with 96.81\% mean accuracy, followed very closely by CNN-1D at 96.64\%, then SVM at 94.53\% and Ridge at 93.50\%. Linear Regression remains substantially lower at 63.42\%, with very large error values and variability across folds. For VBMLR, the confusion matrix (Table~\ref{tab:vbmlr_cm_70}) shows that most
errors concentrate in the upper screen zones in the present experimental setup. A plausible explanation is that, at 70~cm, small variations in the relative position of the subject, chair, and laptop had a stronger impact on the extracted depth and pose related features, making some neighboring zones more confusable than at 50~cm. In particular, $AoR$~2 and $AoR$~4 exhibit the strongest mixing with their adjacent neighbors, while the lower $AoR$s (7--9) remain essentially perfect. This pattern supports a geometric interpretation that at longer range, depth-related features, notably $T_z$ and its interaction with $\delta_x,\delta_y$, become noisier with a webcam.

\begin{table}[H]
\caption{Global comparison (50~cm and 70~cm pooled) using only non-pretrained models. Metrics are averaged over the 9 $AoR$ zones (800 samples per zone).}
\centering
\small
\setlength{\tabcolsep}{4pt}
\resizebox{\textwidth}{!}{%
\begin{tabular}{lcccccccc}
\hline
Method & Mean Acc.\ (\%) & Mean FP & Mean TN & Training Time (ms) & Equivalent Params & Min Err.\ (\%) & Max Err.\ (\%) & Std Err.\ (\%) \\
\hline
VBMLR (proposed) & 87.03 & 101.44 & 6298.56 & 0.0019 & \textbf{81} & 12.15 & 13.61 & 0.49 \\
SVM (RBF)        & 92.79 & 64.89  & 6335.11 & 0.4716 & 35{,}073 & 6.53  & 8.13  & 0.52 \\
LR Multiclass    & 54.61 & 363.11 & 6036.89 & \textbf{0.0003} & \textbf{81} & 44.17 & 47.08 & 1.05 \\
Ridge            & 86.26 & 108.78 & 6291.22 & 0.0063 & 324 & 13.33 & 14.38 & 0.42 \\
CNN-1D           & \textbf{93.72} & \textbf{49.22}  & \textbf{6350.78} & 0.0071 & 2{,}809 & \textbf{6.04}  & \textbf{6.67}  & \textbf{0.23} \\
\hline
\end{tabular}%
}
\label{tab:metrics_global}
\end{table}

To assess robustness when training a single shared model on pooled 50~cm and 70~cm samples, we also report a global evaluation. The aim of this experiment is to produce a model that learns from multiple people and multiple distances. This setting is substantially more challenging because each $AoR$ zone must now be recognized under two distinct depth conditions using a single shared model. As shown in Table~\ref{tab:metrics_global}, VBMLR drops to 87.03\%, with a minimum fold error of 12.15\%, a maximum fold error of 13.61\%, and a standard deviation of 0.49\%, indicating a clear loss of robustness to distance variability in the pooled setting. Ridge follows a similar trend and reaches 86.26\%, with even larger error values. By contrast, SVM (92.79\%) and CNN-1D (93.72\%) degrade more gracefully, with CNN-1D achieving the strongest overall performance among the non-pretrained models, together with the lowest FP count, the highest TN count, and the tightest error statistics. The VBMLR confusion matrix in Table~\ref{tab:vbmlr_cm_global} indicates that the dominant confusions are concentrated among the upper $AoR$s (1--6), while the lower $AoR$s (8--9) remain highly reliable, with recall values above 97\%. One plausible explanation is that a single shared model trained on both 50~cm and 70~cm samples is more affected by the combined variability across distances and subjects, which makes the separation between these neighboring zones less stable. This result suggests that the current VBMLR formulation does not model multimodality strongly enough across viewing distances, since the depth-related feature $T_z$, in interaction with the other features, appears insufficient to separate the pooled geometric regimes. A natural extension would therefore be to incorporate an RBF kernel, as in SVM, in order to better model the nonlinear separation across viewing distances.

\begin{figure}[!htbp]
    \centering
    \fbox{\includegraphics[width=0.92\textwidth]{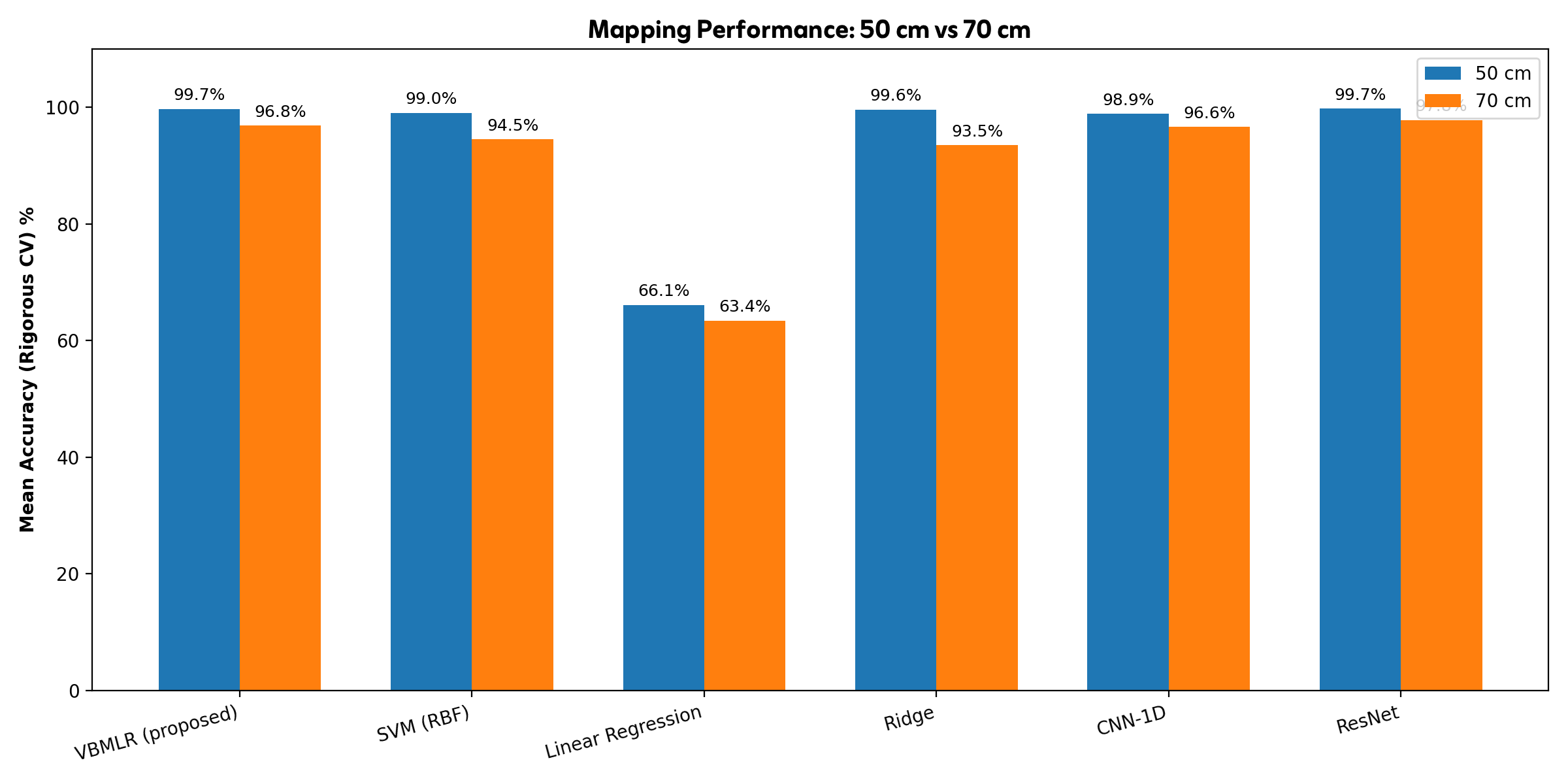}}
    \caption{Mean zone accuracy (\%) at 50~cm and 70~cm for all evaluated mappings under 5-fold stratified cross-validation.}
    \label{fig:mappings_comparison_50_70}
\end{figure}

To resume, figure~\ref{fig:mappings_comparison_50_70} reports the mean zone accuracy (\%) at 50~cm and 70~cm for all compared methods.

In practice, these results support the design choice of VBMLR as our primary method. First, it achieves near-perfect performance for all single-session and fixed working distance with extremely low inference time (about $0.002$ ms/sample) and needs only 81 parameters. Second, it remains lightweight and interpretable as VBMLR requires no GPU and outputs calibrated posterior probabilities rather than opaque
scores. This probabilistic output is particularly valuable for uncertainty-aware decision rules. 

Tables~\ref{tab:vbmlr_cm_50}--\ref{tab:vbmlr_cm_global} report the full zone by zone
count tables for the proposed VBMLR method. Entries are sample counts; the \% column is
the per zone recall and the \# column is the number of test samples for that zone.

\begin{table}[H]
\centering
\textbf{VBMLR (Proposed) --- 50~cm}\par\medskip
\resizebox{\textwidth}{!}{
\begin{tabular}{|l||*{11}{c|}}\hline
$AoR$ Classification & 1&2&3&4&5&6&7&8&9&\makebox[3em]{\%}&\makebox[3em]{\#}\\\hline\hline
1 & \textbf{398}&&&2&&&&&& 99.5 & 400\\\hline
2 && \textbf{400}&&&&&&&& 100.0 & 400\\\hline
3 &&& \textbf{397}&&&3&&&& 99.25 & 400\\\hline
4 & 2&&& \textbf{398}&&&&&& 99.5 & 400\\\hline
5 &&&&& \textbf{400}&&&&& 100.0 & 400\\\hline
6 &&&&3&& \textbf{396}&&&1& 99.0 & 400\\\hline
7 &&&&&&& \textbf{400}&&& 100.0 & 400\\\hline
8 &&&&&&&& \textbf{400}&& 100.0 & 400\\\hline
9 &&&&&&&&& \textbf{400}& 100.0 & 400\\\hline
Total &&&&&&&&&& \textbf{99.69} & \textbf{3600}\\\hline
\end{tabular}}
\caption{VBMLR confusion matrix at 50~cm (4 subjects; 400 samples per zone).}
\label{tab:vbmlr_cm_50}
\end{table}

\begin{table}[H]
\centering
\textbf{VBMLR (Proposed) --- 70~cm}\par\medskip
\resizebox{\textwidth}{!}{
\begin{tabular}{|l||*{11}{c|}}\hline
$AoR$ Classification & 1&2&3&4&5&6&7&8&9&\makebox[3em]{\%}&\makebox[3em]{\#}\\\hline\hline
1 & \textbf{387}&1&&12&&&&&& 96.75 & 400\\\hline
2 & 23& \textbf{361}&7&4&5&&&&& 90.25 & 400\\\hline
3 &&1& \textbf{399}&&&&&&& 99.75 & 400\\\hline
4 & 25&4&& \textbf{366}&2&&3&&& 91.50 & 400\\\hline
5 &&5&&12& \textbf{383}&&&&& 95.75 & 400\\\hline
6 &&&8&&3& \textbf{389}&&&& 97.25 & 400\\\hline
7 &&&&&&& \textbf{400}&&& 100.0 & 400\\\hline
8 &&&&&&&& \textbf{400}&& 100.0 & 400\\\hline
9 &&&&&&&&& \textbf{400}& 100.0 & 400\\\hline
Total &&&&&&&&&& \textbf{96.81} & \textbf{3600}\\\hline
\end{tabular}}
\caption{VBMLR confusion matrix at 70~cm (4 subjects; 400 samples per zone). Confusion concentrates on spatially adjacent $AoR$s in the upper screen region.}
\label{tab:vbmlr_cm_70}
\end{table}

\begin{table}[H]
\centering
\textbf{VBMLR (Proposed) --- Global (50~cm $+$ 70~cm)}\par\medskip
\resizebox{\textwidth}{!}{
\begin{tabular}{|l||*{11}{c|}}\hline
$AoR$ Classification & 1&2&3&4&5&6&7&8&9&\makebox[3em]{\%}&\makebox[3em]{\#}\\\hline\hline
1 & \textbf{597}&24&&166&13&&&&& 74.63 & 800\\\hline
2 & 43& \textbf{628}&13&75&41&&&&& 78.50 & 800\\\hline
3 &&2& \textbf{762}&&7&29&&&& 95.25 & 800\\\hline
4 & 106&18&& \textbf{612}&44&&20&&& 76.50 & 800\\\hline
5 & 12&43&25&22& \textbf{686}&5&&7&& 85.75 & 800\\\hline
6 &&&78&&14& \textbf{687}&&&21& 85.88 & 800\\\hline
7 & 7&&&27&14&& \textbf{731}&21&& 91.38 & 800\\\hline
8 &&&&&1&1&10& \textbf{787}&2& 98.38 & 800\\\hline
9 &&&&&&6&&18& \textbf{776}& 97.00 & 800\\\hline
Total &&&&&&&&&& \textbf{87.03} & \textbf{7200}\\\hline
\end{tabular}}
\caption{VBMLR confusion matrix in the global pooled setting (50~cm $+$ 70~cm; 800 samples per zone). Cross-distance confusion concentrates in depth-sensitive upper $AoR$s (1--6); lower $AoR$s (8--9) remain above 97\%.}
\label{tab:vbmlr_cm_global}
\end{table}

\section{Conclusion}

In this paper, we presented an eye-gaze estimation system that requires only a 2D camera such as an integrated webcam, in contrast with many existing gaze-tracking systems that rely on infrared illumination, multiple cameras, or dedicated hardware. By estimating facial relief from blur, the proposed pipeline also reduces the need for external markers or additional tracking components. The experimental results show that the proposed VBMLR-based approach achieves near-perfect accuracy under the evaluated acquisition conditions, with the user positioned 50 cm or 70 cm from the screen, while remaining computationally lightweight, probabilistic, and analytically tractable. The comparison across mappings further clarifies the contribution of the proposed method. In fixed-distance conditions, VBMLR remains competitive compared to five other methods while offering a simpler training and deployment profile than CNN-based alternatives. Overall, the scores support the use of VBMLR for low-cost webcam-based screen interaction scenarios where the embedded camera is fixed or known during a session. They also motivate several targeted extensions for broader deployment, including evaluations under real-world conditions, reducing the error when training on multiple users placed at different distances, and reducing the size of the areas of regard. Future experiments will expand the subject pool and acquisition conditions to better assess inter-subject generalization. A key direction will be cross-distance adaptation. Investigating distance-aware normalization and domain adaptation strategies could preserve the efficiency of VBMLR while extending its robustness to varying user-to-screen geometries.

\section*{Data availability}

Data will be made available on request.

\section*{Declaration of competing interest}

The authors declare that they have no known competing financial interests or personal relationships that could have appeared to influence the work reported in this paper.

\bibliographystyle{ieeetr}
\bibliography{references}

\end{document}